\documentclass{vgtc}                          

\graphicspath{{figures/}{pictures/}{images/}{./}} 

\usepackage{times}                     

\usepackage{tabu}                      
\usepackage{booktabs}                  
\usepackage{lipsum}                    
\usepackage{mwe}                       

\usepackage{mathptmx}                  
\usepackage{amsmath}                   
\usepackage{colortbl}
\usepackage{multirow}
\usepackage{diagbox}
\usepackage{xcolor}

\onlineid{0}

\vgtccategory{Research}

\vgtcinsertpkg

\title{MSGS: Multispectral 3D Gaussian Splatting}

\author{
\begin{tabular}{ccccc}
Iris Zheng\thanks{iris.zheng@vuw.ac.nz} & 
Guojun Tang\thanks{guojun.tang@vuw.ac.nz} & 
Alexander Doronin\thanks{alex.doronin@vuw.ac.nz} & 
Paul Teal\thanks{paul.teal@vuw.ac.nz} & 
Fang-Lue Zhang\thanks{fanglue.zhang@vuw.ac.nz, the corresponding author}
\end{tabular}
\\
Victoria University of Wellington
}

\teaser{
  \centering
  \includegraphics[width=\linewidth]{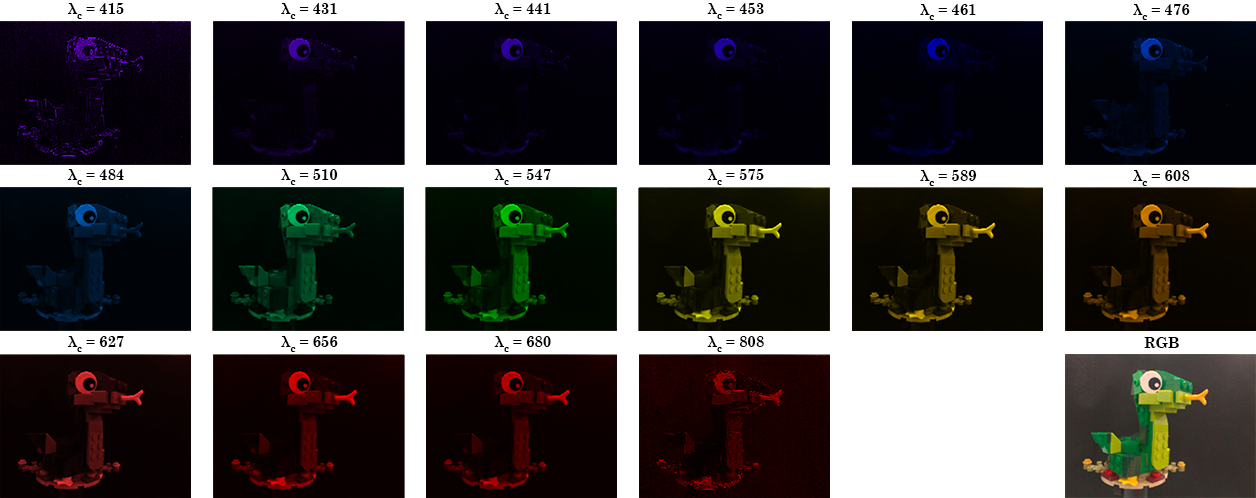}
  \caption{Real-world multispectral (MS) dataset \textit{Snake} captured by our multispectral camera system. $\lambda_c$ denotes the corresponding wavelength. The images are shown for visualization purposes: the original reflectance values (ranging from 0 to 1) in each MS grayscale image are mapped to RGB space for display. The final RGB image is obtained by converting multiple MS grayscale images into RGB using~\autoref{eq:xyz_conversion}.}
  \label{fig:teaser}
}

\abstract{
    We present a multispectral extension to 3D Gaussian Splatting (3DGS) for wavelength-aware view synthesis. Each Gaussian is augmented with spectral radiance, represented via per-band spherical harmonics, and optimized under a dual-loss supervision scheme combining RGB and multispectral signals. To improve rendering fidelity, we perform spectral-to-RGB conversion at the pixel level, allowing richer spectral cues to be retained during optimization. Our method is evaluated on both public and self-captured real-world datasets, demonstrating consistent improvements over the RGB-only 3DGS baseline in terms of image quality and spectral consistency. Notably, it excels in challenging scenes involving translucent materials and anisotropic reflections. The proposed approach maintains the compactness and real-time efficiency of 3DGS while laying the foundation for future integration with physically based shading models.
} 


\begin{document}


\firstsection{Introduction}

\maketitle

Multispectral information plays a crucial role in a wide range of real-world graphics and vision applications where material properties, lighting conditions, and appearance must be accurately captured and reproduced. Fields such as cultural heritage preservation, biomedical imaging, and realistic AR/VR rendering rely on spectral fidelity beyond the visible RGB channels to perform wavelength-aware analysis and physically plausible rendering. For example, subtle differences in surface reflectance can reveal material composition, enable non-invasive diagnostics, or support realistic rendering under varied illuminants—all of which demand accurate multispectral representations.

Despite recent advances in neural and point-based rendering, most existing learning-based 3D reconstruction methods—including the widely adopted 3D Gaussian Splatting (3DGS)~\cite{kerbl_3d_2023}—are fundamentally limited to RGB inputs. These models are typically optimized for photorealistic novel view synthesis in the visible spectrum but are incapable of modeling wavelength-dependent radiance variations essential for spectral analysis and physically grounded rendering. This restricts their applicability in scenarios where color alone is insufficient to distinguish between visually similar but spectrally distinct materials.

In the original 3DGS pipeline, each Gaussian encodes view-dependent color via low-order spherical harmonics (SH) coefficients across the RGB channels, with color parameters often initialized from structure-from-motion (SfM) pipelines. While this formulation is effective for RGB reconstruction, it lacks the flexibility to represent high-demensional wavelength-dependent radiance distributions, which are essential for multispectral rendering.

\begin{figure*}[t]
    \centering
    \includegraphics[width=0.9\linewidth]{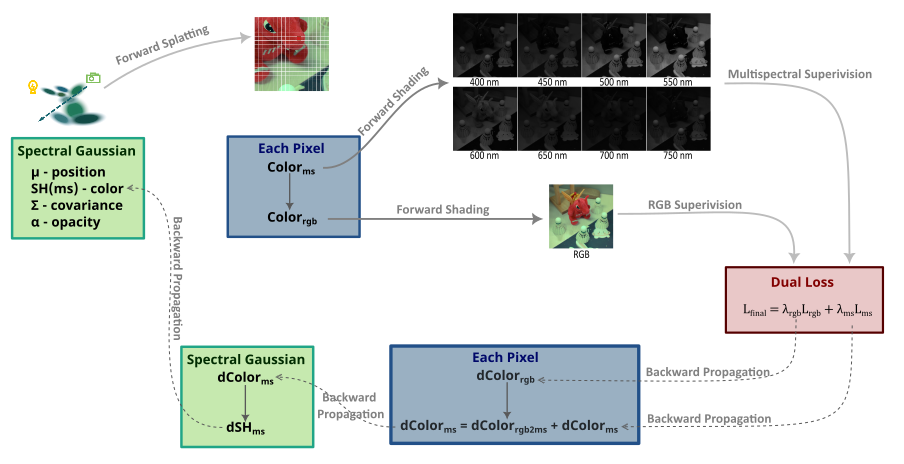}
    \caption{Overview of the proposed Multispectral Gaussian Splatting pipeline. Dual supervision from RGB and multispectral images guides SH parameter updates through both spectral and RGB color spaces.}
    \label{fig:framework}
    \vspace{-0.15in}
\end{figure*}

To address this limitation, we propose a wavelength-aware extension of the 3DGS framework that enables rendering from a broad range of spectral dimensions while preserving its compactness and efficiency. Specifically, we introduce a spectral SH-based radiance representation that generalizes the original RGB encoding, allowing each Gaussian to store and render rich spectral information. In addition, we propose a dual-loss supervision strategy that jointly optimizes multispectral and RGB reconstruction, facilitating robust training. We validate our approach on both public and self-captured MS datasets, demonstrating significant improvements in spectral fidelity and rendering quality over the RGB-only baselines.

\section{Related Work}
Multispectral imaging (MSI) has been widely applied in remote sensing, medical diagnostics, and material analysis, enabling radiance capture across wavelengths beyond the visible RGB range. Traditional MSI systems typically rely on CIE-defined spectral-to-RGB pipelines, involving color matching functions and XYZ-to-sRGB transformations~\cite{wyszecki_color_2000}. Recent learning-based approaches have explored data-driven solutions for multispectral reconstruction and view synthesis~\cite{poggi_cross-spectral_2022, li_spectralnerf_2024, sinha_spectralgaussians_2024, guo_cross-spectral_nodate}.

Neural Radiance Fields (NeRF)~\cite{mildenhall_nerf_2020} have become a dominant framework for novel view synthesis by learning continuous volumetric scene properties via differentiable rendering. Several NeRF variants target the multispectral domain. X-NeRF~\cite{poggi_cross-spectral_2022} addresses cross-spectral synthesis from heterogeneous sensors (e.g., RGB, MS, IR) using joint optimization and a Normalized Cross-Device Coordinate (NXDC) system , rather than improving rendering quality itself. In contrast, SpectralNeRF~\cite{li_spectralnerf_2024} targets physically based spectral rendering by predicting per-wavelength radiance and employing a Spectrum Attention UNet (SAUNet) to fuse outputs into RGB. While flexible, these methods suffer from high computational cost and slow inference due to their reliance on implicit fields and MLP-based rendering.

3D Gaussian Splatting (3DGS)~\cite{kerbl_3d_2023} offers a compelling alternative to NeRF, replacing implicit volumetric fields with explicit 3D Gaussians for real-time rendering and improved geometric clarity. Recent efforts have sought to adapt 3DGS to the multispectral setting. SpectralGaussians~\cite{sinha_spectralgaussians_2024} enhances the original 3DGS by integrating two key components: the Gaussian Shader~\cite{jiang_gaussianshader_2023}, which augments each Gaussian with additional attributes such as surface normals and physically based reflectance (PBR) parameters; and Gaussian Grouping~\cite{ye_gaussian_2023}, which clusters Gaussians based on semantic or material similarity to promote structural coherence and reuse. These enhancements enable more expressive and physically plausible appearance modeling. However, the method creates an independent set of Gaussians for each spectral channel, which breaks geometric consistency across wavelengths and significantly increases memory consumption. Meanwhile, SOC-GS~\cite{guo_cross-spectral_nodate}, addresses cross-spectral pose misalignment by enforcing spatial occupancy consistency across spectra and adopting a two-stage optimization strategy for camera poses and scene representation. Similar in motivation to X-NeRF, SOC-GS focuses primarily on improving alignment rather than enhancing rendering fidelity. It lacks explicit modeling of wavelength-dependent reflectance or support for physically-based spectral shading, limiting its effectiveness in scenes with complex material properties.

\section{Methodology}
We propose a spectral 3D Gaussian Splatting representation where each Gaussian encoded multispectral radiance. We also propose a dual-loss supervision strategy that jointly leverages RGB and multispectral data for high-fidelity scene reconstruction. This design preserves the efficiency and geometric compactness of 3DGS while enabling wavelength-aware rendering. An overview of the framework is shown in~\autoref{fig:framework}.

\begin{figure*}[t]
    \centering
    \includegraphics[width=0.9\linewidth]{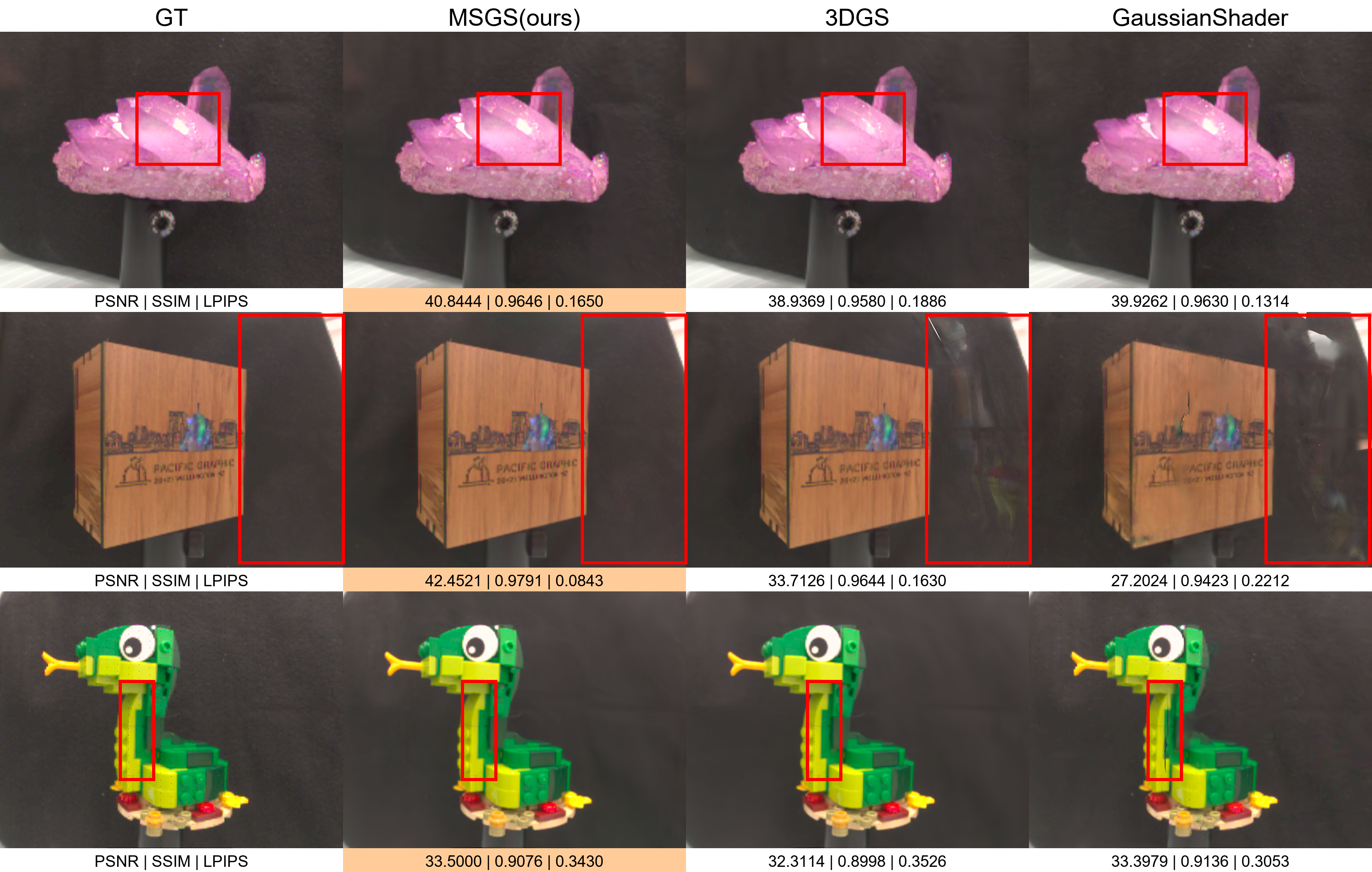}
    \caption{Qualitative Comparison. In the first row, our method better preserves high-frequency reflections compared to 3DGS. In the second row, both 3DGS and Gaussian Shader fails to reconstruct details in dark background regions, while our method maintains structural consistency. In the third row, despite GaussianShader achieving the highest scores on the snake dataset, noticeable artifacts often occur — for example, a long visible crack appears along the snake's neck.}
    \label{fig:qualitative_comparison}
    \vspace{-0.15in}
\end{figure*}

\subsection{Spectral-Aware Gaussian}
We generalize the radiance representation of each Gaussian from a 3-channel RGB spherical harmonics (SH)~\cite{ramamoorthi_efficient_2001, sloan_precomputed_2002} coefficients to an $N$-channel multispectral Spherical-Harmonics (MS-SH) coefficients, where $N$ denotes the number of wavelength bands captured by the imaging system. Specifically, the zeroth-order MS-SH coefficients are initialized from the base colors extracted via COLMAP~\cite{schonberger_structure--motion_2016}, while higher-order terms are zero-initialized and optimized during training to capture angular variations. Meanwhile, each Gaussian retains a single set of geometric attributes—position, scale, and rotation—shared across all spectral channels. This ensures spectral coherence while avoiding redundant memory usage.

Before the forward rendering stage, we first convert the MS-SH coefficients associated with each Gaussian into its corresponding spectral values, following the approach used in vanilla 3DGS~\cite{kerbl_3d_2023}. Based on these spectral values, we perform alpha blending along each camera ray using the standard Gaussian Splatting compositing model, formulated as:

\vspace{-0.2in}
\begin{equation}
\mathbf{C}(p) = \sum_{i=1}^{N} \mathbf{c}_i \cdot \alpha_i(p) \cdot \prod_{j=1}^{i-1} \left(1 - \alpha_j(p)\right),
\label{eq:alpha_blending}
\end{equation}
where overlapping Gaussians contribute to the final pixel value according to their learned opacity and 2D projected footprint. This process produces a per-pixel spectral value map, which serves as the input for the subsequent spectral-to-RGB color conversion stage.

Following the deferred shading concept~\cite{deering_triangle_1988} and its recent adapation to 3DGS~\cite{ye_3d_2024}, we convert the per-pixel multispectral color into RGB color using CIE 1931~\cite{wyszecki_color_2000} color matching functions~\cite{devlin_tone_2002}. Specifically, the spectral radiance $L(\lambda)$  is converted to tristimulus values $(X, Y, Z)$ as:

\vspace{-0.2in}
\begin{equation}
    X = \sum_{\lambda} L(\lambda) \bar{x}(\lambda), \quad
    Y = \sum_{\lambda} L(\lambda) \bar{y}(\lambda), \quad
    Z = \sum_{\lambda} L(\lambda) \bar{z}(\lambda)
    \label{eq:xyz_conversion}
    \vspace{-0.1in}
\end{equation}
where $\bar{x}(\lambda)$, $\bar{y}(\lambda)$, $\bar{z}(\lambda)$ represent the CIE 1931 standard observer functions sampled at the input wavelengths.

To further ensure perceptual fidelity, we apply D65 white-balance correction, aligning the rendering with daylight illumination. Gamma correction is then applied to map the linear XYZ values to display-friendly RGB values.

\subsection{Dual-loss Supervision}
\label{sec:dual_loss}
To effectively train the proposed multispectral 3D Gaussian Splatting model, we introduce a dual-loss strategy that jointly optimizes both multispectral accuracy and RGB rendering consistency. This is achieved by combining two complementary loss terms: a multispectral reconstruction loss $\mathcal{L}_{\mathrm{MS}}$, computed directly between rendered and ground-truth spectral data, and an RGB reconstruction loss $\mathcal{L}_{\mathrm{RGB}}$, computed after converting the spectral outputs to RGB.

A critical design decision lies in choosing the appropriate stage for spectral-to-RGB conversion, as it directly affects gradient propagation and the fidelity of reconstruction. We explore two strategies: the Gaussian-level conversion, which transforms each Gaussian's spectral radiance to RGB before rasterization, and the pixel-level conversion, which defers this conversion until after pixel accumulation. While the former simplifies RGB supervision, it discards spectral details early in the pipeline. In contrast, the latter retains full spectral resolution throughout rendering and supervision, leading to better reconstruction quality.

The final objective combines both loss terms using weighted coefficients:
\begin{equation}
    \mathcal{L}_{\text{total}} = \lambda_{\text{MS}} \mathcal{L}_{\text{MS}} + \lambda_{\text{RGB}} \mathcal{L}_{\text{RGB}}
    \label{eq:total_loss}
    \vspace{-0.1in}
\end{equation}
where $\lambda_{\text{MS}}$ and $\lambda_{\text{RGB}}$ balance the relative contributions of each loss component.
\label{sec:supplement_inst}

For a more detailed evaluation of the dual-loss design and conversion strategy, please refer to~\autoref{sec:ablation_study}.
\begin{table}[h]
    \centering
    \caption{Quantitative comparison with the original 3DGS~\cite{kerbl_3d_2023} and PBR-based Gaussian Shader~\cite{jiang_gaussianshader_2023}. All three methods are trained for 30K iterations under identical resolution and supervision settings. Our method achieves better scores on most datasets. The best results are colored in \colorbox[rgb]{1,0.8,0.8}{\strut red}.}
    \label{tab:quantitative}
    \vspace{-0.1in}
    
    \resizebox{\linewidth}{!}{
    \begin{tabular}{l|ccccccc}
        \hline
        \diagbox{Method}{Dataset} & Projector & Dragon & Onion & Mushroom & Snake & Crystal & Box \\
        \hline
        \multicolumn{8}{c}{\textbf{PSNR} $\uparrow$} \\
        \hline
        3DGS & 36.7580 & 38.1470 & 37.8868 & {\cellcolor[rgb]{1,0.8,0.8}}34.5161 & 33.0125 & 37.4461 & 37.9612 \\
        GaussianShader & 30.0921 & 34.4725 & 35.8232 & 32.8344 & {\cellcolor[rgb]{1,0.8,0.8}}34.3955 & 37.6846 & 34.1192 \\
        Ours & {\cellcolor[rgb]{1,0.8,0.8}}36.8304 & {\cellcolor[rgb]{1,0.8,0.8}}38.5589 & {\cellcolor[rgb]{1,0.8,0.8}}38.3313 & 33.9437 & 33.8852 & {\cellcolor[rgb]{1,0.8,0.8}}37.6949 & {\cellcolor[rgb]{1,0.8,0.8}}39.9916 \\
        \hline
        \multicolumn{8}{c}{\textbf{SSIM} $\uparrow$} \\
        \hline
        3DGS & 0.9607 & 0.9715 & 0.9254 & 0.9139 & 0.9042 & 0.9571 & 0.9615 \\
        GaussianShader & 0.9210 & 0.9550 & {\cellcolor[rgb]{1,0.8,0.8}}0.9308 & 0.9098 & {\cellcolor[rgb]{1,0.8,0.8}}0.9221 & 0.9576 & 0.9508 \\
        Ours & {\cellcolor[rgb]{1,0.8,0.8}}0.9619 & {\cellcolor[rgb]{1,0.8,0.8}}0.9717 & 0.9293 & {\cellcolor[rgb]{1,0.8,0.8}}0.9233 & 0.9121 & {\cellcolor[rgb]{1,0.8,0.8}}0.9634 & {\cellcolor[rgb]{1,0.8,0.8}}0.9686 \\
        \hline
        \multicolumn{8}{c}{\textbf{LPIPS} $\downarrow$} \\
        \hline
        3DGS & 0.0967 & {\cellcolor[rgb]{1,0.8,0.8}}0.0745 & 0.1634 & 0.2214 & 0.3478 & 0.1852 & 0.1538 \\
        GaussianShader & 0.1911 & 0.1149 & 0.1515 & 0.2582 & {\cellcolor[rgb]{1,0.8,0.8}}0.2818 & 0.1642 & 0.1911 \\
        Ours & {\cellcolor[rgb]{1,0.8,0.8}}0.0915 & 0.0747 & {\cellcolor[rgb]{1,0.8,0.8}}0.1491 & {\cellcolor[rgb]{1,0.8,0.8}}0.2093 & 0.3392 & {\cellcolor[rgb]{1,0.8,0.8}}0.1628 & {\cellcolor[rgb]{1,0.8,0.8}}0.1302 \\
        \hline
    \end{tabular}
    }
    
\vspace{-0.2in}
\end{table}

\section{Results}
We evaluate our method on both public and self-captured multispectral data to validate its effectiveness in spectral-aware NVS and RGB rendering quality. Results are reported using both quantitative metrics and qualitative comparisons. Additionally, we conduct ablation studies to analyze the impact of our dual-loss supervision strategy and pixel-level spectral-to-RGB conversion design.

\subsection{Datasets}
Our experiments are conducted on a combination of public real-world multispectral datasets and self-captured data. Specifically, we use two real-world scenes from the SpectralNeRF~\cite{li_spectralnerf_2024} dataset—\textit{Dragon} and \textit{Project}—captured with RGB cameras and rotating filter wheels. Each scene contains images sequentially captured across eight narrow spectral bands spanning 400–750\,nm.

To further evaluate generalizability across diverse material properties and geometries, we capture five additional scenes—\textit{Onion}, \textit{Mushroom}, \textit{Snake}, \textit{Crystal}, and \textit{Box}—using a multispectral camera system (S1-EVK2b~\cite{noauthor_s1-evk2b_nodate}) that directly acquires spectral images across 16 bands in the 415–808,nm range, as illustrated in~\autoref{fig:teaser}. These scenes were carefully selected to cover a wide range of material and appearance characteristics. \textit{Onion} and \textit{Mushroom} represent smooth, homogeneous diffuse surfaces with low-frequency texture, ideal for evaluating color consistency and shading. \textit{Snake} is a colorful LEGO object composed of high-frequency patterns and synthetic materials. \textit{Crystal} features strong translucency and partial transparency, challenging both geometry and appearance reconstruction. \textit{Box} is a wooden container made from abalone shell, with complex, view-dependent reflectance due to its iridescent surface. All datasets provide per-pixel spectral measurements along with corresponding RGB images, enabling both spectral and RGB supervision for joint evaluation.

\subsection{Quantitative and Qualitative Analysis}
Although recent works like X-NeRF~\cite{poggi_cross-spectral_2022}, SpectralNeRF~\cite{li_spectralnerf_2024}, SpectralGaussians~\cite{sinha_spectralgaussians_2024}, and SOC-GS~\cite{guo_cross-spectral_nodate} have explored multispectral view synthesis, none provide official implementations for direct comparison. We therefore compare against the original 3DGS baseline~\cite{kerbl_3d_2023} and PBR-based Gaussian Shader~\cite{jiang_gaussianshader_2023}, focusing on RGB reconstruction accuracy. Our method consistently outperforms 3DGS in most scenes (see~\autoref{tab:quantitative}).

Qualitative comparisons are provided for both RGB and multispectral outputs: for RGB, we visually compare our results with those from 3DGS and Gaussian Shader, as presented in~\autoref{fig:qualitative_comparison}; for multispectral outputs, we showcase rendered spectral images produced by our model, as 3DGS cannot represent or synthesize spectral data. Notably, \textit{Crystal} presents a challenging case due to its translucent material and complex optical effects, while \textit{Box} contains an anisotropic reflective icon that highlights view-dependent appearance variations. Our method demonstrates improved fidelity and robustness in both cases.

\begin{table}[h]
    \centering
    \caption{Comparison between conversion strategies at Gaussian and Pixel Level across 7K and 30K iterations. The results are evaluated on our self-captured \textit{crystal} dataset. The best/second-best results are colored in \colorbox[rgb]{1,0.8,0.8}{\strut red} / \colorbox[rgb]{1,0.9,0.8}{\strut orange}, respectively.}
    \label{tab:ablation_study_conversion-strategies}
    \vspace{-0.1in}
    
    \resizebox{\linewidth}{!}{
        \begin{tabular}{cc|cccc|cccc}
            \hline
            \multicolumn{2}{c|}{\multirow{2}{*}{\diagbox{Method}{Iteration}}} 
            & \multicolumn{4}{c|}{7000} & \multicolumn{4}{c}{30000} \\
            \cline{3-10}
            & & Time $\downarrow$ & PSNR $\uparrow$ & SSIM $\uparrow$ & LPIPS $\downarrow$ 
              & Time $\downarrow$ & PSNR $\uparrow$ & SSIM $\uparrow$ & LPIPS $\downarrow$ \\
            \hline
            \multirow{3}{*}{\shortstack{Conversion \\at Gaussian \\ Level}} & RGB & {\cellcolor[rgb]{1,0.9,0.8}}04'40 & 34.0308 & 0.9505 & 0.2074 & {\cellcolor[rgb]{1,0.9,0.8}}22'08 & {\cellcolor[rgb]{1,0.9,0.8}}37.2594 & {\cellcolor[rgb]{1,0.9,0.8}}0.9572 & {\cellcolor[rgb]{1,0.9,0.8}}0.1866 \\
            & MS & 06'39 & 29.1552 & 0.9352 & {\cellcolor[rgb]{1,0.9,0.8}}0.2295 & 29'04 & 29.3512 & 0.9387 & 0.2200 \\
            & MS \& RGB & 09'47 & 29.0900 & 0.9374 & 0.2276 & 54'01 & 26.9797 & 0.9382 & 0.2315 \\
            \hline
            \multirow{3}{*}{\shortstack{Conversion \\at Pixel \\ Level}} & RGB & {\cellcolor[rgb]{1,0.8,0.8}}04'23 & 33.9366 & {\cellcolor[rgb]{1,0.9,0.8}}0.9505 & 0.2072 & {\cellcolor[rgb]{1,0.8,0.8}}18'54 & 36.9047 & 0.9569 & 0.1877 \\
            & MS & 07'05 & {\cellcolor[rgb]{1,0.9,0.8}}34.0480 & 0.9472 & 0.2145 & 30'40 & 36.6371 & 0.9528 & 0.1984 \\
            & MS \& RGB & 09'00 & {\cellcolor[rgb]{1,0.8,0.8}}34.4375 & {\cellcolor[rgb]{1,0.8,0.8}}0.9543 & {\cellcolor[rgb]{1,0.8,0.8}}0.1936 & 41'56 & {\cellcolor[rgb]{1,0.8,0.8}}38.6076 & {\cellcolor[rgb]{1,0.8,0.8}}0.9624 & {\cellcolor[rgb]{1,0.8,0.8}}0.1662 \\
            \hline
        \end{tabular}
    }
\end{table}

\vspace{-0.2in}
\subsection{Ablation Study}
\label{sec:ablation_study}
We perform ablation studies on three key components: (1) the dual-loss supervision strategy, (2) the level of spectral-to-RGB conversion, and (3) the choice of spectral band range. See~\autoref{tab:ablation_study_conversion-strategies} and~\autoref{tab:ablation_study_spectral-band} for detailed results.

\paragraph{1. Loss Supervision Strategy.}
We evaluate three loss settings: RGB-only loss, MS-only loss, and full dual-loss. The dual-loss strategy consistently achieves the best performance across different scenes and iteration checkpoints (7K and 30K), providing a balanced trade-off between perceptual quality and spectral accuracy. RGB-only supervision lacks spectral fidelity, while MS-only training, though spectrally accurate, can result in perceptual color distortions due to the absence of visual priors.

\paragraph{2. Spectral-to-RGB Conversion Strategy.}
We further compare two different strategies for spectral-to-RGB conversion: one at the Gaussian level (before rasterization), and the other at the pixel level (after rasterization). Results show that pixel-level conversion consistently yields better performance. Interestingly, when applying dual-loss supervision with Gaussian-level conversion, the performance at 30K iterations is even worse than at 7K, and lower than using either loss alone. This suggests that converting spectral data to RGB before rasterization—an inherently nonlinear projection—may introduce conflicting gradient directions during backpropagation, especially after blending in the rasterization step. Such conflicts could hinder optimization and reduce convergence stability.

\paragraph{3. Spectral Band Selection.}
To investigate the difference between visible and near-infrared performance, we conduct experiments on four band configurations using our self-captured \textit{crystal} dataset. The 808\,nm band, which lies in the near-infrared region beyond the visible spectrum, provides valuable structural information and contributes positively to reconstruction quality. Interestingly, however, the full spectral range from 415 to 808\,nm does not achieve the best overall performance. In fact, the 415–680\,nm configuration yields the lowest scores across all metrics, suggesting that the 415\,nm band—despite being within the visible range—may introduce excessive noise due to sensor limitations, thus degrading the results in both cases where it is included. Among the remaining settings, the 431–808\,nm range achieves the highest PSNR, while the 431–680\,nm range performs better in terms of SSIM and LPIPS. This indicates a trade-off: near-infrared bands improve structural accuracy, whereas visible-only bands offer better perceptual quality, depending on the rendering objective.

\begin{table}[h]
    \centering
    \caption{Comparison of different spectral band selections using our self-captured \textit{crystal} dataset across 7K and 30K iterations. The best/second-best results are colored in \colorbox[rgb]{1,0.8,0.8}{\strut red} / \colorbox[rgb]{1,0.9,0.8}{\strut orange}, respectively.}
    \label{tab:ablation_study_spectral-band}
    \vspace{-0.1in}

    \resizebox{\linewidth}{!}{
        \begin{tabular}{c|cccc|cccc}
            \hline
            \multirow{2}{*}{\diagbox[width=3.5cm]{Band Range}{Iteration}} & \multicolumn{4}{c|}{7K Iterations} & \multicolumn{4}{c}{30K Iterations} \\
            \cline{2-9}
            & Time $\downarrow$ & PSNR $\uparrow$ & SSIM $\uparrow$ & LPIPS $\downarrow$ 
            & Time $\downarrow$ & PSNR $\uparrow$ & SSIM $\uparrow$ & LPIPS $\downarrow$ \\
            \hline
            431–680\,nm & {\cellcolor[rgb]{1,0.8,0.8}}13'45 & {\cellcolor[rgb]{1,0.9,0.8}}34.8798 & {\cellcolor[rgb]{1,0.8,0.8}}0.9542 & {\cellcolor[rgb]{1,0.8,0.8}}0.1949 & {\cellcolor[rgb]{1,0.8,0.8}}1:07'07 & 38.1892 & {\cellcolor[rgb]{1,0.8,0.8}}0.9610 & {\cellcolor[rgb]{1,0.8,0.8}}0.1739 \\
            415–680\,nm & {\cellcolor[rgb]{1,0.9,0.8}}14'51 & 34.5260 & 0.9522 & {\cellcolor[rgb]{1,0.9,0.8}}0.1986 & 1:12'28 & 37.5136 & 0.9588 & 0.1778 \\
            431–808\,nm & 14'58 & {\cellcolor[rgb]{1,0.8,0.8}}35.2441 & {\cellcolor[rgb]{1,0.9,0.8}}0.9528 & 0.1993 & 1:12'22 & {\cellcolor[rgb]{1,0.8,0.8}}38.6689 & {\cellcolor[rgb]{1,0.9,0.8}}0.9598 & {\cellcolor[rgb]{1,0.9,0.8}}0.1765 \\
            415–808\,nm & 15'01 & 34.8456 & 0.9507 & 0.2037 & {\cellcolor[rgb]{1,0.9,0.8}}1:10'41 & {\cellcolor[rgb]{1,0.9,0.8}}38.5316 & 0.9586 & 0.1797 \\
            \hline
        \end{tabular}
    }
\end{table}

\vspace{-0.2in}
\section{Conclusions and future work}
We proposed a multispectral extension to 3D Gaussian Splatting (3DGS) that enables wavelength-aware view synthesis while maintaining the efficiency and geometric compactness of the original framework. By augmenting the color representation with spectral SH-based radiance and introducing a dual-loss supervision strategy, our method incorporates multispectral information without duplicating geometry and achieves superior rendering performance across both RGB and spectral domains. Experimental results highlight the benefits of leveraging richer spectral data, especially in scenes with complex materials or lighting conditions.

For future work, we aim to improve the physical accuracy of spectral radiance representation to achieve higher visual fidelity in NVS. While the current SH-based model~\cite{ramamoorthi_efficient_2001, sloan_precomputed_2002} is efficient, it remains limited in capturing anisotropic reflectance and fine angular details. To address this, future extensions may adopt more expressive basis functions such as ASG~\cite{xu_anisotropic_2013}, or incorporate physically grounded models like BRDFs and BSSRDFs for better handling of complex light transport. Additionally, learning adaptive weights for multispectral channels and losses could further improve optimization and generalization.

\acknowledgments{
This work was supported by Marsden Fund Council managed by the Royal Society of New Zealand under Grant MFP-20-VUW-180. The authors would like to thank Xuan Zhu and Simin Kou for their helpful discussions and valuable suggestions during the development of this work. }

\bibliographystyle{abbrv-doi}

\bibliography{ISMAR-2025}
\end{document}